\documentclass{article} 
\usepackage{iclr2026_conference,times}

\usepackage{amsmath,amsfonts,bm}

\def\eqref#1{equation~\ref{#1}}

\def\1{\bm{1}}

\DeclareMathAlphabet{\mathsfit}{\encodingdefault}{\sfdefault}{m}{sl}
\SetMathAlphabet{\mathsfit}{bold}{\encodingdefault}{\sfdefault}{bx}{n}

\usepackage{hyperref}
\usepackage{url}
\usepackage{graphicx}
\usepackage{subcaption}
\usepackage{wrapfig}
\usepackage{booktabs}

\title{Mitigating Compounding Error via Video Representation Regularization}

\author{Taiye Chen, Qi Zhang, Yisen Wang\thanks{Corresponding author: Yisen Wang (yisen.wang@pku.edu.cn)} \\
Peking University\\
}

\iclrfinalcopy 
\begin{document}

\maketitle

\begin{abstract}
Video diffusion-based world models enable long autoregressive video generation for robotics, autonomous driving and simulation tasks, yet sliding-window autoregressive inference suffers from severe error accumulation that degrades frame quality over time. Although this phenomenon has been widely observed, the underlying mechanism of compounding error and how to achieve stable long-horizon generation remain largely unresolved. In this paper, we investigate the internal representation dynamics of video world models and discover that compounding error is tightly coupled with dimensional collapse of hidden representations. Specifically, the effective rank of model representations sharply decreases at the onset of generation drift, revealing a strong connection between representational degradation and long-term rollout instability. Furthermore, we find that pure training data scaling fails to boost model resistance to error drift, contradicting mainstream scaling paradigms. To address this problem, we propose video representation regularization, a lightweight training constraint that stabilizes latent representations and suppresses iterative error accumulation. Compared with Diffusion Forcing, our method achieves improvements from 38.65 to 55.56 and from 44.37 to 72.08 on the Aesthetic Quality and Imaging Quality metrics of VBench. Our work establishes the first connection between autoregressive video drifting and model internal representations, adopts erank as a quantitative metric for error accumulation, reveals counterintuitive scaling limitations for video world models, and presents a simple yet effective regularization strategy to improve long video generation robustness.
\end{abstract}

\section{Introduction}

World models have recently attracted significant attention from the research community and technology companies due to their advantages in areas such as data synthesis, model-based planning, and simulation. Among these, driven by the continuous advancement of video generation technology, video world models have demonstrated tremendous influence and promising prospects in domains including autonomous driving~\citep{hu2023gaia, ren2025cosmosdrivedreamsscalablesyntheticdriving}, navigation~\citep{bar2024navigation}, robotic manipulation~\citep{wu2024ivideogpt, azzolini2025cosmos, ge2025, maes2026leworldmodel}, and games~\citep{valevski2024diffusion, oasis2024, che2024gamegen, guo2025mineworld, yu2025gamefactory}. Video diffusion models, in particular, have become an important implementation pathway for video world models owing to their superior capabilities in high-resolution settings. Some works~\citep{chen2024diffusion,huang2026self,jin2025pyramidal} have further extended them into autoregressive diffusion models, making the generation of infinitely long videos possible.

However, due to the high-dimensional nature of video data, video world models struggle to achieve context lengths on the order of millions of tokens as easily as text-based models do. In contrast, a context consisting of merely tens of frames is sufficient to exhaust the GPU memory of a video world model. To tackle this limitation, the sliding window strategy is widely adopted: previously generated frames act as conditioning inputs for the autoregressive generation of subsequent frames. Nevertheless, these generated frames are inherently imperfect, which leads to rapid deterioration in video quality after tens of iterative generation steps, a phenomenon termed error accumulation, also referred to as drift, exposure bias, or compounding error. While numerous existing studies have proposed diverse techniques to alleviate this issue, a unified and adequately effective solution is still lacking. Worse still, we find that merely expanding the training dataset fails to mitigate compounding errors; instead, it may aggravate the problem. 

By examining videos exhibiting severe error accumulation, we observe that generation often deteriorates into random noise or overexposed frames, indicating a potential degradation of the model’s internal representations. Motivated by this observation, we study how representations evolve throughout autoregressive video generation. We quantify the expressiveness of hidden states using effective rank, which captures the degree of dimensional collapse in learned representations. Interestingly, we find that the decline of effective rank closely coincides with the onset of visual collapse, revealing a strong connection between representation degradation and compounding error. Moreover, as the amount of training data increases, effective rank does not improve and even decreases, suggesting that simply scaling data is insufficient to enhance robustness against error accumulation.

Based on the observations above, we introduce \textbf{video representation regularization (VRR)}, which aims to improve the model's resistance to error accumulation by regularizing the model's representational capacity during training. We are the first to establish a connection between the error accumulation phenomenon in autoregressive video generation and model representations. We argue that as the amount of training data increases, the model tends to learn shortcuts, leading to the collapse of its representations. By incorporating regularization during training, the model can learn better representations, thereby improving its resistance to error accumulation. On VBench, our method achieves substantial improvements over Diffusion Forcing, increasing Aesthetic Quality from 38.65 to 55.56 and Imaging Quality from 44.37 to 72.08. In summary, the main contributions of this paper are as follows:

\begin{itemize}
    \item We identify the relationship between video frame quality degradation and model representations. By introducing representation erank into video generation models, we obtain a sound explanation for frame collapse, which also serves as an effective quantitative metric for error accumulation.
    \item We discover that simply increasing training data fails to improve the model's resistance to error accumulation, which contradicts the conventional wisdom of scaling and provides a valuable reference for both the academic and industrial communities.
    \item We propose video representation regularization, which enhances the model's resistance to error accumulation through the addition of a simple regularization training term.
\end{itemize}

\begin{figure}[t]
    \centering
    \includegraphics[width=1.0\textwidth]{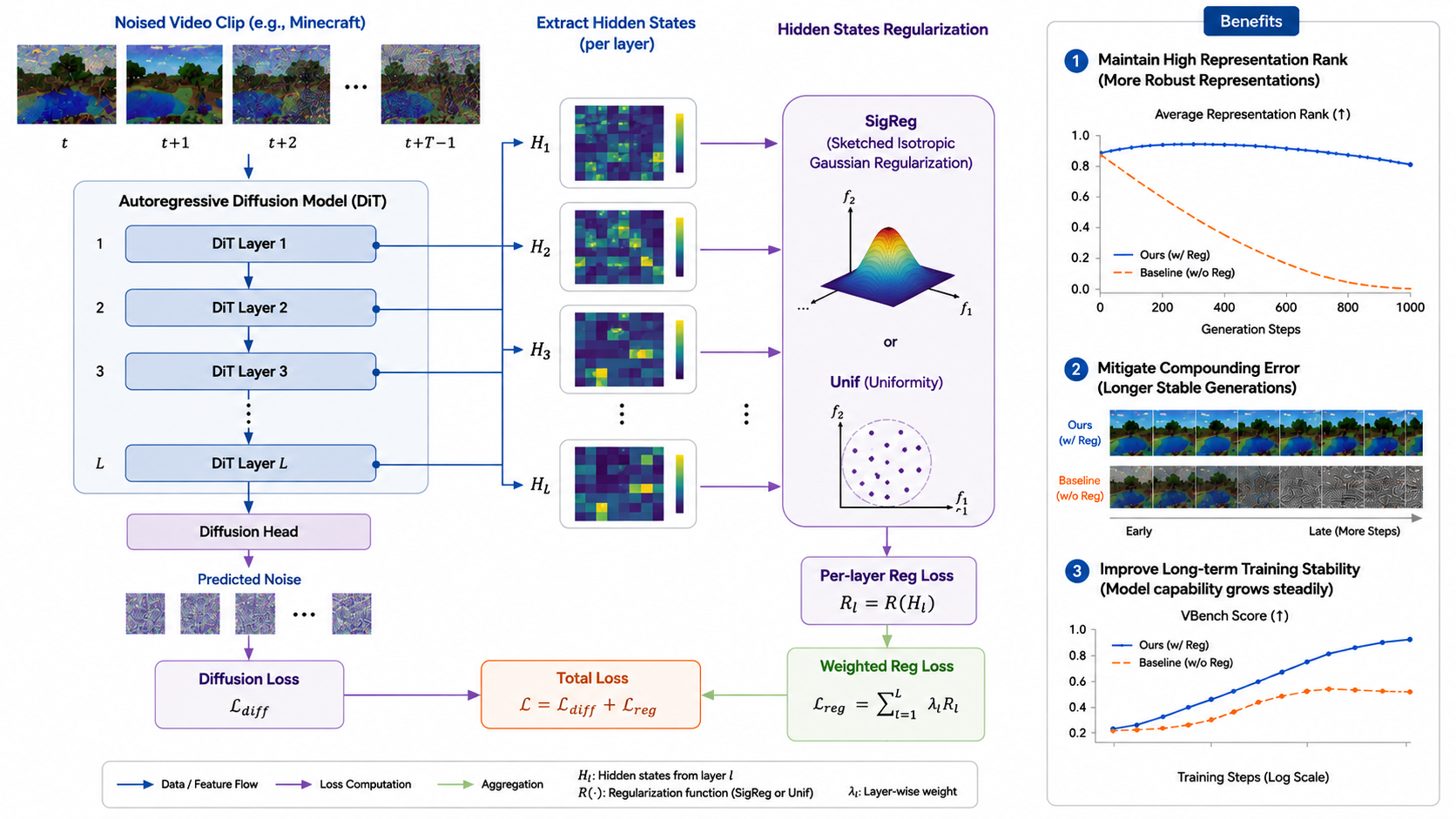}
    \caption{Overview of our proposed method. Existing autoregressive diffusion models suffer from compounding errors during long-range generation, and this issue cannot be effectively mitigated by simply scaling up training data. We introduce hidden states regularization into DiT layers to strengthen the model’s representation capacity, alleviate compounding errors, and enable consistent performance gains as training proceeds.}
    \label{fig:overview}
\end{figure}

\section{Related work}

\subsection{Video Diffusion Model}

Diffusion models have become a dominant paradigm for image and video generation due to their strong sample quality and stable training dynamics~\citep{blattmann2023align, harvey2022flexible, 10377444, blattmann2023stable, chen2024videocrafter2, ho2022video, singer2022make, hong2022cogvideo, yang2024cogvideox, wang2023modelscope, ding2024dollar, opensora}. Early diffusion generators typically adopt UNet-based denoising backbones~\citep{ho2022video, blattmann2023stable, chen2024videocrafter2}, while recent models increasingly replace them with diffusion transformers (DiT)~\citep{peebles2023scalable} to improve scalability and capture long-range spatio-temporal dependencies. To reduce the computational cost of high-resolution generation, many systems perform denoising in the latent space of a variational autoencoder (VAE)~\citep{kingma2013auto}, rather than directly in pixel space~\citep{rombach2022high, blattmann2023stable}. However, most existing video diffusion models are trained to generate clips with a fixed number of frames. To extend generation beyond the training horizon, several methods use sliding-window or overlapping-context inference, producing long videos by repeatedly denoising local temporal windows~\citep{jin2025pyramidal,chen2024diffusion,Yin_2025_CVPR,cui2025self,xie2024progressive}.

\subsection{Video World Model and Long Video Generation}

World models aim to predict future states conditioned on past observations and, when available, actions~\citep{ha2018world}. Existing approaches differ in the space where prediction is performed. Representation-based methods, such as JEPA-style models, first encode visual observations into latent representations and then learn predictive dynamics in that feature space~\citep{bardes2024revisitingfeaturepredictionlearning,assran2025vjepa2selfsupervisedvideo,zhou2025dinowmworldmodelspretrained,maes2026leworldmodel}. In contrast, generative world models use video generation models to directly predict future visual observations, often at the pixel or latent-pixel level~\citep{wu2024ivideogpt,chen2026learning,parkerholder2024genie2,genie3,nvidia2025cosmosworldfoundationmodel}. This paper focuses on the latter setting, where a video diffusion model is rolled out auto-regressively for long-horizon generation.

Auto-regressive video generation naturally supports variable-length prediction, but it also suffers from compounding error: small mistakes in early generated frames are fed back as conditioning inputs and can gradually accumulate over time~\citep{zhang2025packinginputframecontext,song2025historyguidedvideodiffusion,blattmann2023stable}. This phenomenon is closely related to exposure bias and temporal drifting in sequence generation. Prior work has attempted to mitigate this issue through modified sampling schedules~\citep{qiu2024freenoise,3692070}, frame anchoring~\citep{khachatryan2023text2video,weng2024art}, and training-time noise injection or error recycle~\citep{li2026stable,po2026bagger}. Despite these improvements, maintaining stable long-horizon dynamics remains challenging, especially when the generated frames also serve as the model's future conditioning context.

{{\subsection{Representation helps generation}
Recent studies have highlighted the importance of representations in improving generative models. One line of work introduces semantic representations to guide generative training, where pretrained encoders provide high-level supervision to improve generation quality and training efficiency~\citep{edunov2019pre,NEURIPS2024_e304d374,peale2025representative}. For instance, REPA aligns diffusion features with representations from pretrained visual encoders, encouraging diffusion models to learn more semantic representations~\citep{yu2024repa}, while recent representation autoencoder (RAE)-based approaches further demonstrate the benefits of expressive latent representations for diffusion models \citep{rae}. Beyond representation alignment, representation regularization has also been explored to preserve feature quality and training stability. DispLoss~\citep{wang2025diffusedisperseimagegeneration} introduces representation-based objectives into the denoising process to accelerate convergence and improve sample fidelity. In video world models, LeJEPA adopts SigReg~\citep{balestriero2025lejepa} as a regularizer to prevent latent representation collapse and maintain informative dynamics. However, existing studies mainly focus on improving representation quality during training, while the role of representation degradation in long-horizon autoregressive rollout remains unexplored.
}

\section{Core Observations: Error Accumulation Coupled with Representation Collapse and Limits of Data Scaling}

\subsection{Preliminary: Auto-regressive Video Diffusion}

\paragraph{Video Diffusion Model}
Video diffusion models operate either directly on pixel-level inputs $\boldsymbol{x}\in \mathbb{R}^{L \times H \times W \times 3}$, or on a latent representation $\boldsymbol{z} = \mathcal{E}(\boldsymbol{x})$, where $\mathcal{E}(\cdot)$ is a pretrained variational autoencoder (VAE) encoder. The forward process gradually adds Gaussian noise to the input according to a variance schedule $\{\beta_t\}_{t=1}^T$:
\begin{equation}
    q(\boldsymbol{z}_t|\boldsymbol{z}_{t-1}) = \mathcal{N}(\boldsymbol{z}_t; \sqrt{1-\beta_t}\boldsymbol{z}_{t-1}, \beta_t\mathbf{I})
\end{equation}

The model learns to reverse this process by predicting the noise $\boldsymbol{\epsilon}_\theta$ at each step:
\begin{equation}
    \mathcal{L} = \mathbb{E}_{t,\boldsymbol{\epsilon},\boldsymbol{z}}\left[\|\boldsymbol{\epsilon} - \boldsymbol{\epsilon}_\theta(\boldsymbol{z}_t, t)\|_2^2\right]
\end{equation}
where $\boldsymbol{z}_t = \sqrt{\bar{\alpha}_t}\,\boldsymbol{z}_0 + \sqrt{1-\bar{\alpha}_t}\,\boldsymbol{\epsilon}$ with $\boldsymbol{\epsilon} \sim \mathcal{N}(\mathbf{0},\mathbf{I})$.

At inference time, new videos are generated by starting from random noise $\boldsymbol{z}_T \sim \mathcal{N}(\mathbf{0},\mathbf{I})$ and iteratively denoising:
\begin{equation}
    \boldsymbol{z}_{t-1} = \frac{1}{\sqrt{\alpha_t}}\!\left(\boldsymbol{z}_t - \frac{\beta_t}{\sqrt{1-\bar{\alpha}_t}}\boldsymbol{\epsilon}_\theta(\boldsymbol{z}_t,t)\right) + \sigma_t\boldsymbol{\epsilon}
\end{equation}
where $\alpha_t = 1-\beta_t$ and $\bar{\alpha}_t = \prod_{s=1}^t \alpha_s$.

\paragraph{Diffusion Forcing}
To enable long video generation, we apply the Diffusion Forcing~\citep{chen2024diffusion} technique. During training, we randomly add noise to each frame in the entire input video sequence according to the diffusion schedule: $z^i_t = \sqrt{\bar{\alpha}_t} z^i_0 + \sqrt{1-\bar{\alpha}_t} \epsilon^i, \epsilon^i \sim \mathcal{N}(0,\mathbf{I})$, where $z^i_t$ represents the noised latent of the $i$-th frame, and the training objective for action-conditioned autoregressive video models become:
\begin{align*}
    \mathcal{L}_\text{DF} &= \mathbb{E}_{[t],\boldsymbol{\epsilon},\mathbf{z},a}[\|\boldsymbol{\epsilon} - \boldsymbol{\epsilon}_\theta(\mathbf{z}_{[t]}, [t], \boldsymbol{a})\|_2^2] \\ \boldsymbol{\epsilon}&=\{\epsilon^i\}_{i=1}^L, \mathbf{z}_{[t]}=\{z^i_t\}_{i=1}^L
\end{align*}
where $[t]$ is vector of $L$ timesteps with different $t\in[T]$ for each frame and $\boldsymbol{a}$ is an action sequence $\boldsymbol{a} \in \mathbb{R}^{L \times A}$. The noise prediction model $\boldsymbol{\epsilon}_\theta$ conditioned on both the action sequence $\boldsymbol{a}$ and noised frames $\mathbf{z}_{[t]}$.

\subsection{Error Accumulation Is Coupled with Representation Collapse}

\begin{figure}[h]
    \centering
    \includegraphics[width=1.0\textwidth]{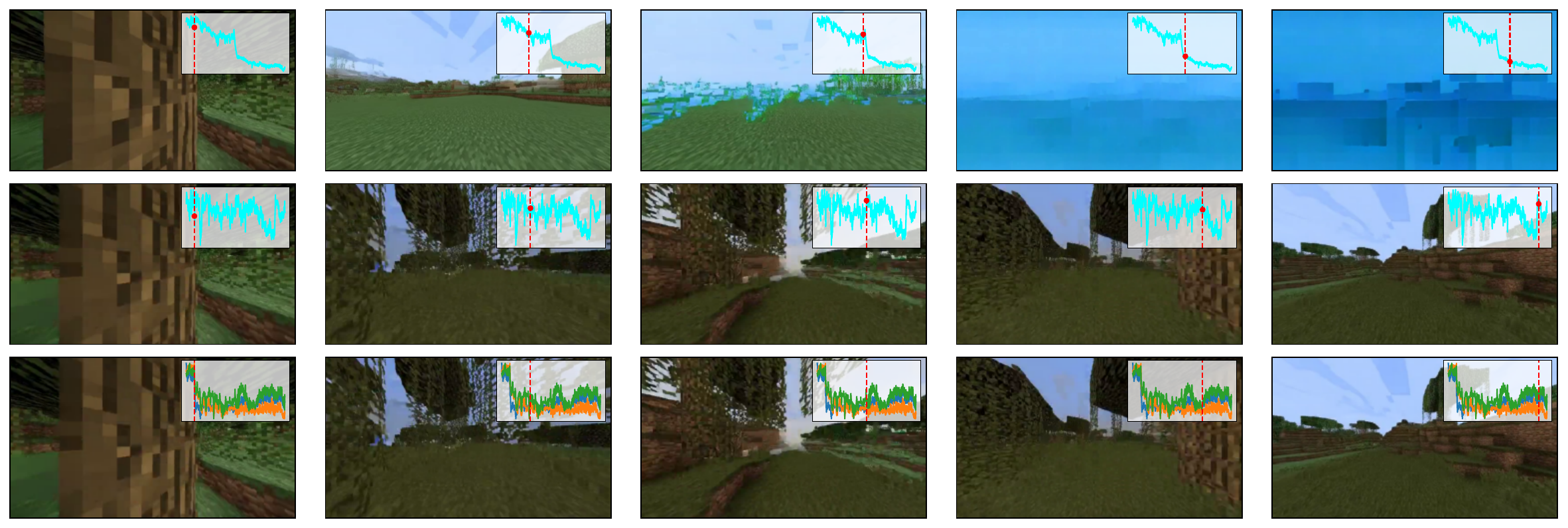}
    \caption{\textbf{Top row:} Video frames sampled from a 40k-step checkpoint trained with the vanilla diffusion forcing method. Hidden states from DiT’s 7th layer are extracted to analyze the effective rank (Erank), which reveals a direct correlation between visual collapse and abrupt Erank reduction. \textbf{Middle row:} Samples generated by a 4k-step checkpoint under the same training scheme; stable Erank values are observed when video content remains coherent. \textbf{Bottom row:} Identical experimental settings as the middle row, with frame quality quantified via normalized SSIM, PSNR and LPIPS.}
    \label{fig:real_video}
\end{figure}

A central finding of our work is that the abrupt onset of error accumulation during autoregressive video generation is intrinsically coupled with representation collapse in the underlying diffusion model.
To quantify the fidelity of internal representations, we measure the \textbf{effective rank}~\citep{roy2007effective} of the hidden states extracted from intermediate layers of the DiT backbone, which captures how many linearly independent directions are actively used by the model at each generation step.

As illustrated in Figure~\ref{fig:real_video}, there is a correspondence between the frame at which generated video quality suddenly drifts—manifesting as semantic incoherence or visual artifacts—and the frame at which the effective rank undergoes a sharp collapse.
This suggests that the model's internal representation loses its expressive capacity precisely when error accumulation becomes catastrophic.

To validate that effective rank is a uniquely informative indicator—and not merely one of many correlated signals—we also conduct a comparison against a suite of alternative diagnostics, including standard image quality metrics such as SSIM, PSNR and LPIPS.
Crucially, none of these alternatives consistently pinpoints the collapse frame.

\subsection{More Data Cannot Cure Error Accumulation}\label{sec:more_data}

Another central finding of our investigation is that naïvely scaling up the training set does not alleviate error accumulation in long video generation.
Following the experimental protocol of VRag~\citep{chen2026learning}, we curate a dataset of 18{,}000 Minecraft gameplay sequences, each spanning 1{,}200 frames.
We train a single epoch on this dataset—deliberately avoiding multiple passes to rule out overfitting—and save checkpoints at regular intervals throughout training.
For each checkpoint we perform long-video inference and measure the effective rank of the generated sequences as a proxy for temporal diversity and structural coherence.

\begin{figure}[h]
    \centering
    \includegraphics[width=1.0\textwidth]{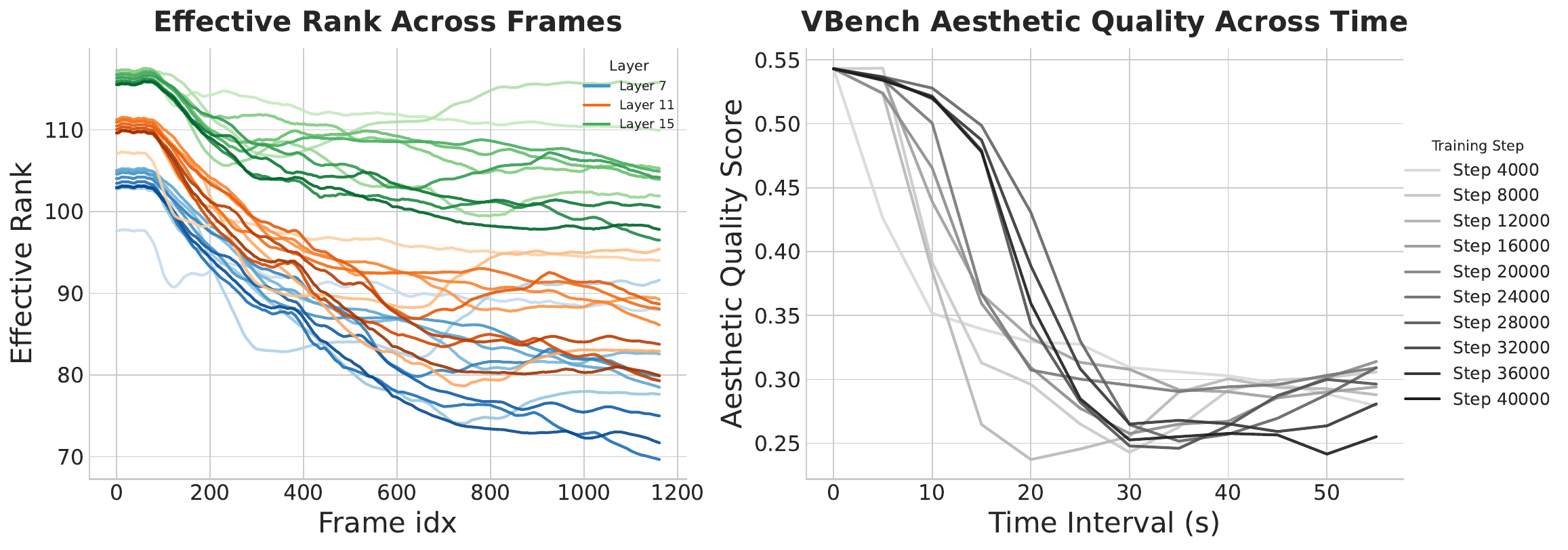}
    \caption{We trained for one epoch on the 18k training set using the vanilla diffusion forcing method and calculated the Erank corresponding to these checkpoints. We observed that the Erank values at step 4000 and step 8000 are generally higher and maintain stability over longer durations. In contrast, increasing the number of training steps not only reduces the overall Erank but also accelerates its degradation during long-range reasoning.}
    \label{fig:erank}
\end{figure}

The results are shown in Figure~\ref{fig:erank}.
Strikingly, the effective rank peaks after exposure to only $\sim$10\% of the training data; every subsequent checkpoint exhibits a monotonic decline.
The ability to leverage large-scale data for improved generalization has long been regarded as a defining strength of video generation models.
Our experiment challenges this assumption in the context of autoregressive generation: error accumulation is not a data-scarcity problem that can be remedied by collecting more trajectories.
Instead, it reflects a structural limitation of the generation paradigm itself, motivating the architectural intervention we introduce in the following section.

\section{Video Representation Regularization}

\subsection{Effective Rank Degradation in Long-term Training}

Motivated by the observations in the previous section, our core objective is to improve
the effective rank (erank) of the diffusion model over long-term training.
Given a hidden state matrix $\mathbf{H}\in\mathbb{R}^{n\times d}$, erank is defined as
\begin{equation}
    \text{erank}(\mathbf{H})
    = \exp\!\left(-\sum_{i=1}^{r} p_i \log p_i\right), \qquad
    p_i = \frac{\sigma_i(\mathbf{H})}{\displaystyle\sum_{j=1}^{r}\sigma_j(\mathbf{H})},
    \label{eq:erank}
\end{equation}
where $\sigma_i(\mathbf{H})$ denotes the $i$-th singular value of $\mathbf{H}$ and $r$ is its rank.
To maintain a high erank throughout training, we impose a representation regularization loss
on the hidden states, encouraging a more uniform singular value distribution.

\subsection{Formulation of Video Representation Regularization}

We argue that the anomalous decreasing trend of effective rank (erank) as training steps increase stems from shortcut learning in autoregressive diffusion models. When frame contents change mildly, directly copying adjacent frames enables smooth temporal consistency and favorable generation quality. Nevertheless, such frame-copying behavior erodes the amount of valid information contained within latent representations.

To preserve representation robustness and deter the model from over-relying on trivial shortcuts, a straightforward solution is to introduce regularization. To this end, we propose Video Representation Regularization (VRR). Given an n-layer DiT model \(D\), let \(x\) denote a video clip input with frame-wise independent noise levels and \(y\) denote its corresponding ground-truth video clip. Let \(H_1, ..., H_n\) represent the hidden states from each layer. The overall training objective is formulated as:
\[
\mathcal{L} = \mathcal{L}_{\text{DF}}(D(x), y) + \sum_i \lambda_i \mathcal{L}_{\text{reg}}(H_i)
\]
where \(\mathcal{L}_{\text{DF}}\) stands for the vanilla Diffusion Forcing loss, \(\lambda_i\) denotes the regularization weight coefficient, and \(\mathcal{L}_{\text{reg}}\) refers to the regularization function.


In this work, we primarily adopt Sigreg and Uniformity loss as our regularization functions. Sigreg regularizes by constraining vectors to an isotropic Gaussian, whereas Uniformity Loss regularizes by constraining vectors to follow a uniform distribution. We also experiment with classic representation regularization terms including Barlow Twins and VICReg. Additionally, we conduct ablation studies that directly utilize effective rank as the regularization function. Detailed experimental results are presented in Section~\ref{sec:abla_reg}.


\section{Experiments}

\subsection{Experimental Setup}\label{sec:setup}

\paragraph{Datasets}

For Minecraft experiments, we use MineRL~\citep{guss2019minerl} to generate 18{,}000 training sequences following the protocols in VRAG~\citep{chen2026learning}. Each video contains 1{,}200 frames. For evaluation, we reserve 20 sequences from the dataset, which are sufficiently long to assess error accumulation in the diffusion model. During inference, we feed the model 100 frames of ground-truth video as the prompt, and the model generates the subsequent 1100 frames. Across all experiments, the window size of the DiT model is set to 20 frames.

\paragraph{Evaluation} We adopt VBench~\citep{10657096} as our evaluation metric. VBench covers multiple assessment dimensions including text-to-video alignment, human characters, object motion, and background consistency, none of which are relevant to the Minecraft setting. For this reason, we primarily refer to two dimensions: imaging quality and aesthetic quality. These two metrics mainly evaluate whether the frames contain noise or suffer from visual collapse.

\paragraph{Baseline} In addition to Diffusion Forcing, we also implement two baselines, namely Frame Anchor and SVI~\citep{li2026stable}. Frame Anchor has been adopted in numerous studies as a strategy to mitigate compounding error~\citep{khachatryan2023text2video,weng2024art}. Its core mechanism involves inserting one completely clean frame as an anchor into the sliding window. SVI improves the model’s robustness against compounding error by recapturing errors generated during training and incorporating them into the training samples.

\paragraph{Model Architecture}

\begin{wrapfigure}{r}{0.5\textwidth}
    \centering
    \includegraphics[width=1.0\linewidth]{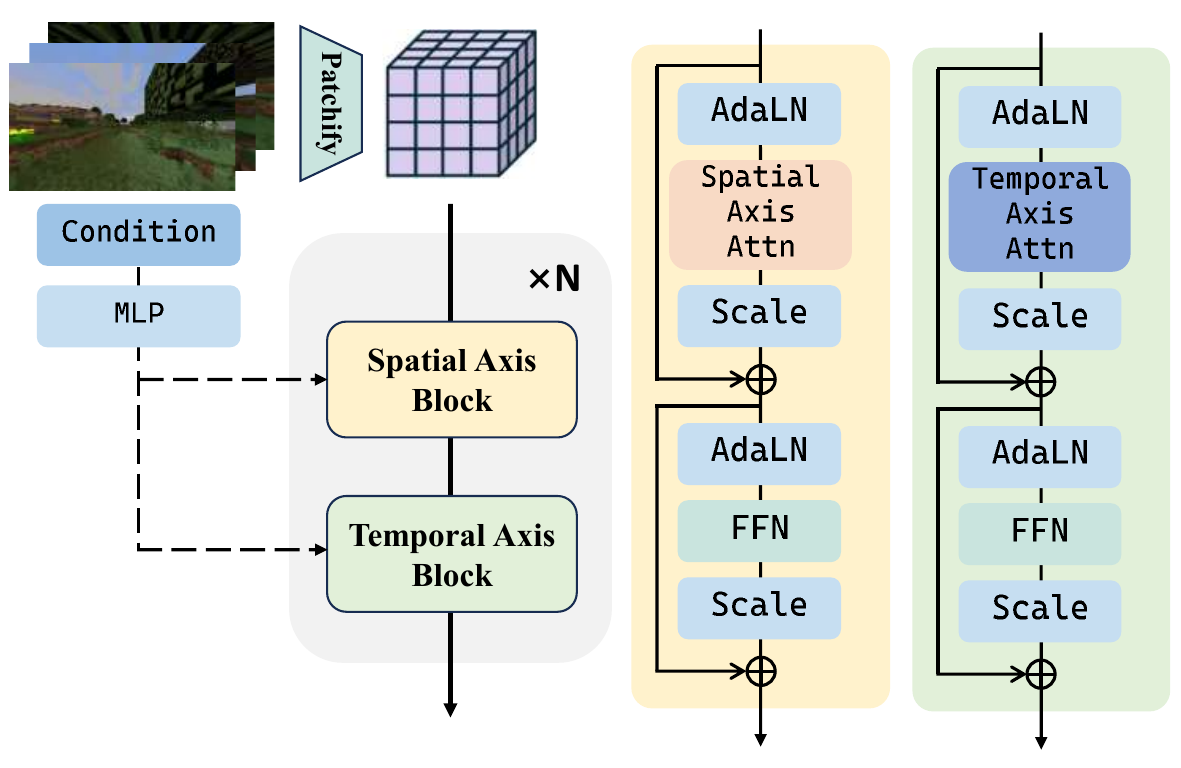}
    \caption{Architecture of DiT model.}
    \label{fig:model_arch}
\end{wrapfigure}

As shown in Figure~\ref{fig:model_arch}, following the designs by previous work~\citep{oasis2024,opensora,chen2026learning}, we decompose the attention mechanism in our DiT into two distinct modules: Spatial Axis Attention and Temporal Axis Attention. The Rotary Position Embedding (RoPE)~\citep{su2024roformer} is applied in both spatial and temporal dimensions, to enhance the capacity of both attention modules to capture spatial positional dependencies and temporal correlations. Conditioning information, specifically the timestep and action condition, is incorporated into the DiT model via adaptive Layer Normalization (adaLN).

\subsection{Experimental Results}

\begin{table}[ht]
    \centering
    \caption{Experimental results of all methods with different checkpoints after 16,000 training steps evaluated on VBench metrics. Our proposed VRR method achieves steady performance improvements throughout training, and consistently outperforms all baseline methods by a considerable margin.}
    \begin{tabular}{l|cccc|cccc}
        \toprule
        VBench Metric & \multicolumn{4}{|c|}{Aesthetic Quality$\uparrow$} & \multicolumn{4}{|c}{Imaging Quality$\uparrow$}\\
        \midrule
        Training Steps & 4000 & 8000 & 12000 & 16000 & 4000 & 8000 & 12000 & 16000 \\
        \midrule
        Diffusion Forcing & 39.57 & 37.52 & 36.82 & 38.65 & 46.20 & 33.78 & 39.46 & 44.37\\
        Frame Archor & \textbf{45.95} &38.13&36.62&39.15&48.32&33.99&40.15&45.38\\
        SVI & 35.62 &36.24&36.35&39.17&43.28&44.82&36.53&42.87\\
        \textbf{VRR-Unif} &41.01&46.09&51.44&53.78&38.84&58.77&\textbf{70.54}&\textbf{72.08} \\
        \textbf{VRR-Sigreg} & 44.60 & \textbf{52.24} &\textbf{54.46}&\textbf{55.56}&\textbf{60.92} & \textbf{67.87} & 68.87 & 69.51\\
        \bottomrule
    \end{tabular}
    \label{tab:exp_result}
\end{table}

We train all compared methods for 16k training steps under identical experimental setups, and evaluate the holistic quality of several 1-minute generated videos via the VBench benchmark. Quantitative results are summarized in Table~\ref{tab:exp_result}.
For Vanilla Diffusion Forcing, the measured metrics corroborate our observations in Section~\ref{sec:more_data}. Its internal latent representations degrade progressively as training proceeds, which yields a matching declining trend across all VBench scores. Both Aesthetic Quality and Imaging Quality peak at merely 4,000 training steps; further prolonged training only degrades the final generation quality.
Except for achieving a relatively high score at the 4000th step, the Frame Anchor method performs comparably to the diffusion forcing baseline with only marginal improvements. Meanwhile, SVI is not tailored for autoregressive diffusion architectures, making it ineffective at mitigating compounding prediction errors during sequential generation.
In stark contrast, our representation regularization (VRR) framework achieves outstanding performance. On one hand, our approach avoids performance degradation with extended training iterations, demonstrating strong training robustness. On the other hand, VRR substantially outperforms all baseline methods across all VBench metrics, verifying its superior generation efficiency.

\begin{figure}[t]
    \centering
    \includegraphics[width=1.0\textwidth]{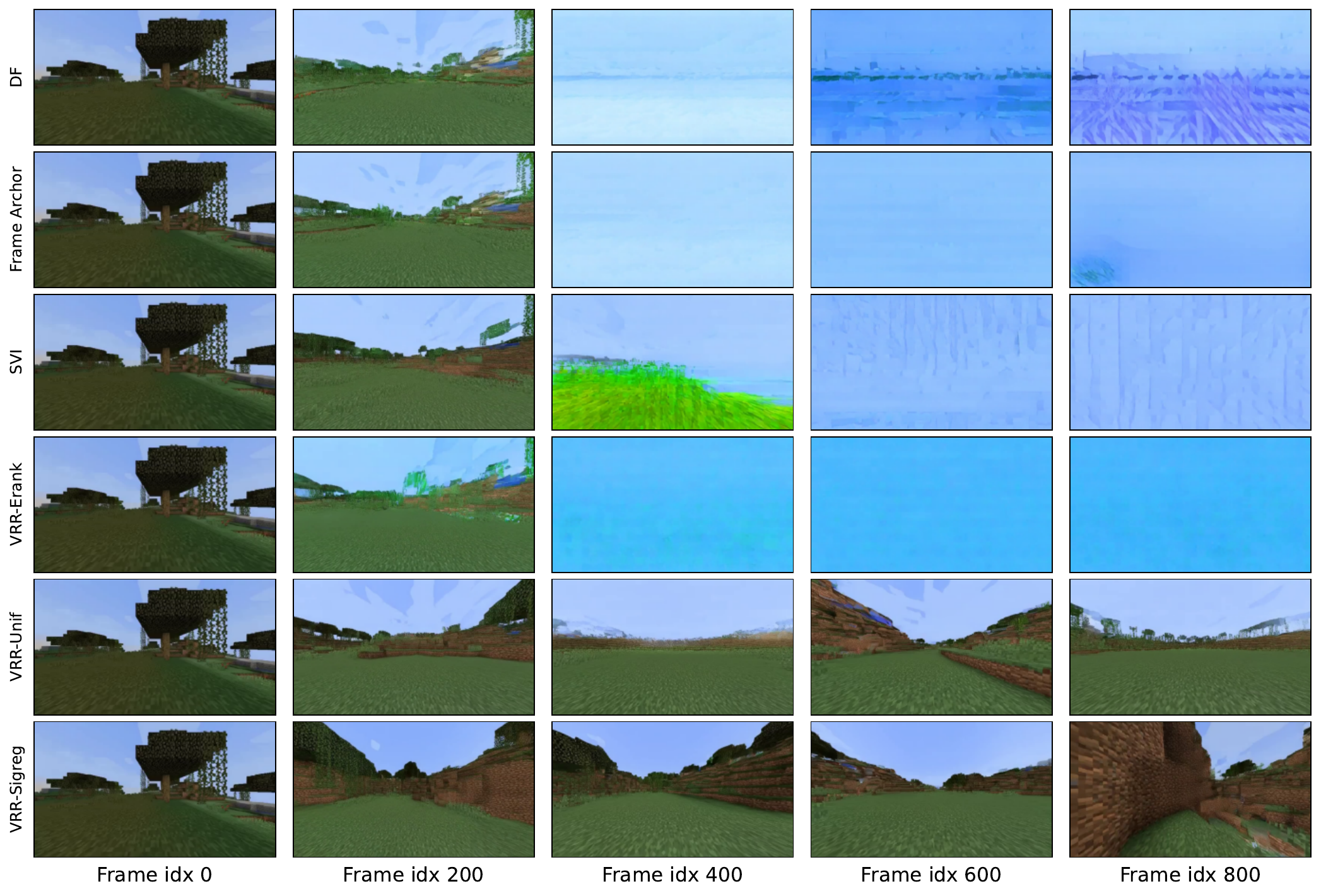}
    \caption{Frames of videos generated from the checkpoint of all methods at the 16,000th training step}
    \label{fig:vis_results}
\end{figure}

To further demonstrate the superiority of VRR, we present qualitative video visualizations in Figure~\ref{fig:vis_results}. All methods take a 100-frame video prompt as the initial condition to perform autoregressive video extension. We observe that all approaches can generate coherent frames after the first 100 frames. Nevertheless, after an additional 200 frames, Diffusion Forcing, Frame Anchor, SVI and VRR-Erank suffer severely from compounding errors and fail to produce plausible video outputs, with subsequent frames collapsing completely. In contrast, our proposed VRR method, whether regularized via Sigreg or Uniformity loss, maintains stable generation over extremely long contexts and is largely immune to the adverse effects of compounding errors.

\subsection{Mitigating Compounding Error}

\begin{figure}[h]
    \centering
    \begin{minipage}{0.48\textwidth}
        \centering
        \includegraphics[width=\linewidth]{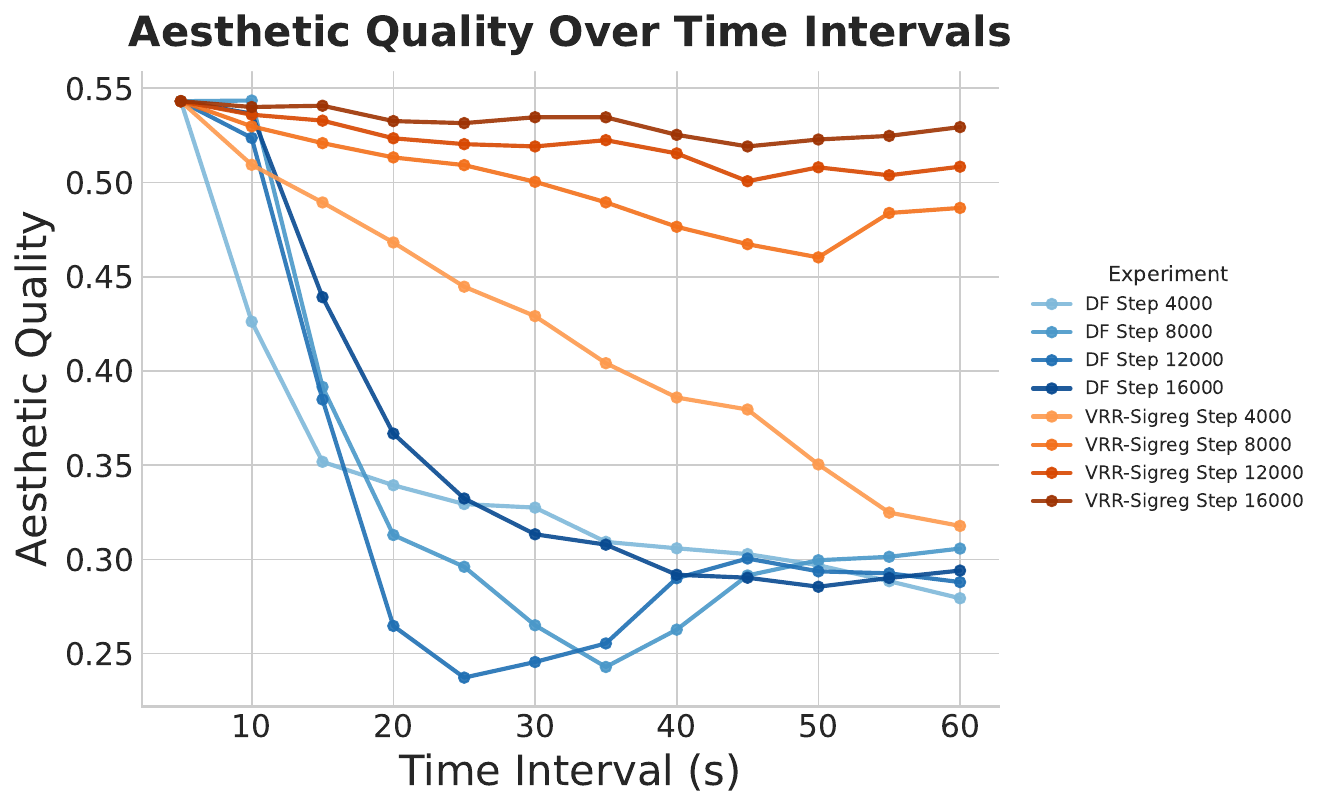}
    \end{minipage}
    \hfill
    \begin{minipage}{0.48\textwidth}
        \centering
        \includegraphics[width=\linewidth]{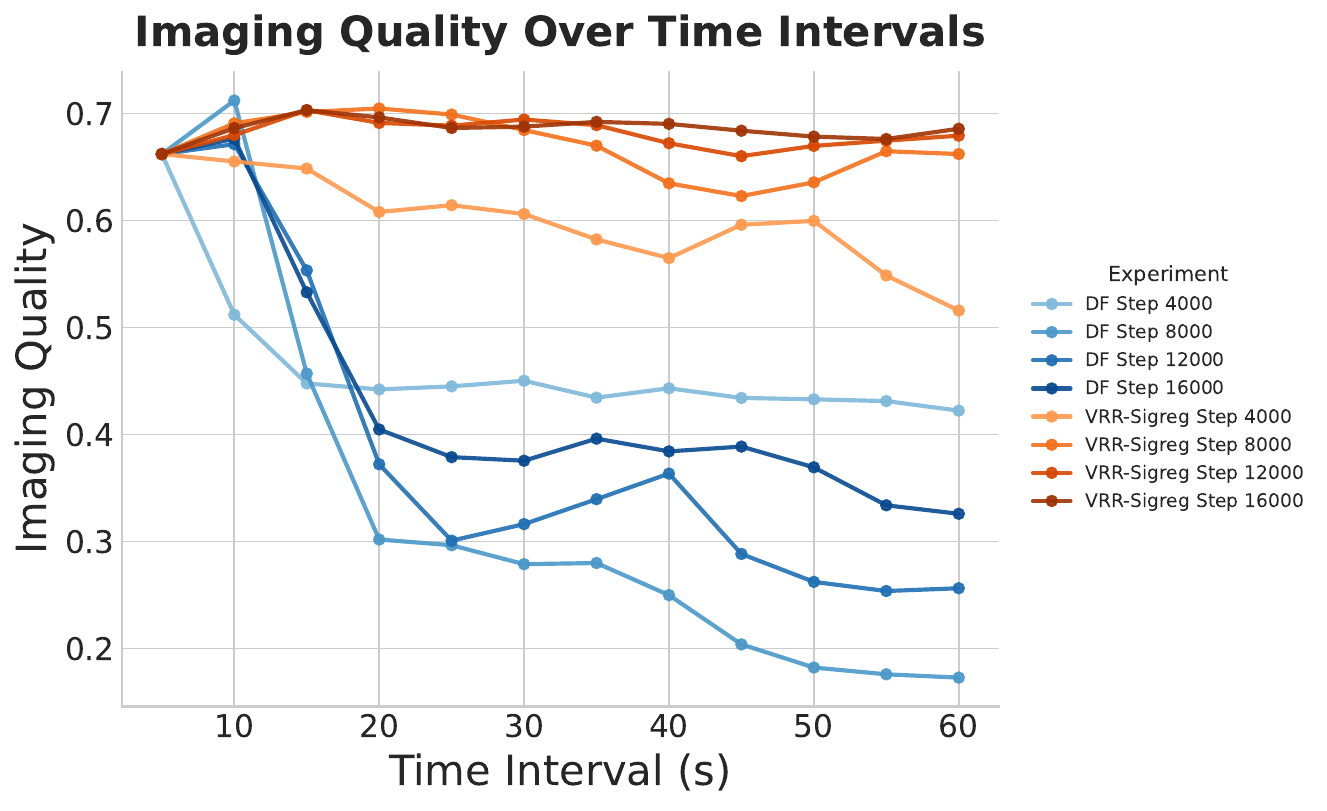}
    \end{minipage}
    \caption{The video quality of vanilla Diffusion Forcing continuously degrades as the video progresses. Our method maintains high scores across VBench metrics.}
    \label{fig:compounding_error_vbench}
\end{figure}

To evaluate whether our method effectively mitigates the compounding error issue, we split each 1-minute long video into 12 non-overlapping 5-second clips, and sequentially evaluate the video quality of each clip using VBench. The results are presented in Figure~\ref{fig:compounding_error_vbench}.
We observe that the video quality of vanilla Diffusion Forcing continuously degrades as the video progresses. In contrast, after incorporating our regularization technique and training for 4000 steps, the compounding error of the model is substantially alleviated, and the model maintains high scores across VBench metrics.

\subsection{Ablation Study}

\paragraph{Comparison of Different Representation Regularization Methods}\label{sec:abla_reg}


\begin{table}[ht]
    \centering
    \caption{Experimental Results of Different Regularization Methods}
    \begin{tabular}{l|cccc|cccc}
        \toprule
        VBench Metric & \multicolumn{4}{|c|}{Aesthetic Quality$\uparrow$} & \multicolumn{4}{|c}{Imaging Quality$\uparrow$}\\
        \midrule
        Training Steps & 4000 & 8000 & 12000 & 16000 & 4000 & 8000 & 12000 & 16000 \\
        \midrule
        Diffusion Forcing & 39.57 & 37.52 & 36.82 & 38.65 & 46.20 & 33.78 & 39.46 & 44.37\\
        \midrule
        Barlow twins &42.56&52.18&38.67&41.44&40.81&49.75&38.76&36.74 \\
        VICReg &34.55&29.37&32.00&41.44&47.93&67.52&46.64&36.74 \\
        Erank &\textbf{45.90}&35.20&39.88&37.05&53.70&36.24&39.30&37.49 \\
        \textbf{Unif} &41.01&46.09&51.44&53.78&38.84&58.77&\textbf{70.54}&\textbf{72.08} \\
        \textbf{Sigreg} & 44.60 & \textbf{52.24} &\textbf{54.46}&\textbf{55.56}&\textbf{60.92} & \textbf{67.87} & 68.87 & 69.51\\
        \bottomrule
    \end{tabular}
    \label{tab:abla_reg}
\end{table}

We apply different representation regularization strategies to the hidden states of the diffusion model and compare their effectiveness.
Results are shown in Table~\ref{tab:abla_reg}.
Both SigReg and Unif achieve comparable performance, which substantially outperforms the baseline. The remaining methods yield mediocre results. The experimental observations of Erank regularization are similar to those of the diffusion forcing baseline. Barlow Twins and VICReg suffer from training instability, which may lead to sharp increases in the diffusion loss.

\paragraph{Effect of Regularizing Hidden States at Different Layers}

\begin{table}[ht]
    \centering
    \caption{Experimental results of regularizing hidden states at different layers using the VRR-Sigreg method}
    \begin{tabular}{l|cccc|cccc}
        \toprule
        VBench Metric & \multicolumn{4}{|c|}{Aesthetic Quality$\uparrow$} & \multicolumn{4}{|c}{Imaging Quality$\uparrow$}\\
        \midrule
        Training Steps & 4000 & 8000 & 12000 & 16000 & 4000 & 8000 & 12000 & 16000 \\
        \midrule
        Diffusion Forcing & 39.57 & 37.52 & 36.82 & 38.65 & 46.20 & 33.78 & 39.46 & 44.37\\
        \midrule
        Reg at layer 7 & 31.99  & \textbf{54.17} & 48.60  &  53.39 & 24.10  & 64.88  &  53.73 & 66.57 \\
        Reg at layer 15 & 36.56 & 32.55 &40.79&38.78 &44.79&46.54&53.98&51.23\\
        
        Reg at all layers & 44.00 &48.21&52.06&51.42& 39.19&47.05&65.82&62.92\\
        Reg at layer 0, 7, 15 & \textbf{44.60} & 52.24 &\textbf{54.46}&\textbf{55.56}&\textbf{60.92} & \textbf{67.87} & \textbf{68.87} & \textbf{69.51} \\
        \bottomrule
    \end{tabular}
    \label{tab:abla_layer}
\end{table}

To investigate which layers of the DiT layer are more critical for mitigating compounding errors, we conducted experiments that apply regularization to the hidden states of different layers. We observed that VRR consistently improves the performance of DiT relative to vanilla diffusion forcing, regardless of the regularization coefficient selected. Nevertheless, applying regularization solely to layers 0, 7, and 15 yields particularly prominent gains. These correspond to the first, middle, and final layers of the DiT model, respectively.

\section{Conclusion and Discussion}

In this work, we present Video Representation Regularization (VRR), a method that regularizes representations during the training of autoregressive diffusion models to boost long-horizon generation performance and mitigate compounding errors. We observe that error accumulation in autoregressive video generation is typically accompanied by dimensional collapse of model representations. Using effective rank as a metric, we verify a strong correlation between the two phenomena. We further find that simply increasing training data cannot effectively alleviate compounding errors. Building on these observations, our VRR method effectively enhances representation expressiveness and reduces compounding errors, achieving substantially better performance than the baselines across VBench metrics. While the efficacy of our approach has been validated, several open questions remain for further investigation: Why does expanding training data lead to degraded representations? What shortcuts does the DiT learn throughout the training process? We leave these directions for future work.

\bibliography{iclr2026_conference}

@article{bar2024navigation,
  title={Navigation world models},
  author={Bar, Amir and Zhou, Gaoyue and Tran, Danny and Darrell, Trevor and LeCun, Yann},
  journal={arXiv preprint arXiv:2412.03572},
  year={2024}
}

@article{guo2025mineworld,
  title={Mineworld: a real-time and open-source interactive world model on minecraft},
  author={Guo, Junliang and Ye, Yang and He, Tianyu and Wu, Haoyu and Jiang, Yushu and Pearce, Tim and Bian, Jiang},
  journal={arXiv preprint arXiv:2504.08388},
  year={2025}
}

@article{opensora,
  title={Open-sora: Democratizing efficient video production for all},
  author={Zheng, Zangwei and Peng, Xiangyu and Yang, Tianji and Shen, Chenhui and Li, Shenggui and Liu, Hongxin and Zhou, Yukun and Li, Tianyi and You, Yang},
  journal={arXiv preprint arXiv:2412.20404},
  year={2024}
}

@article{che2024gamegen,
  title={Gamegen-x: Interactive open-world game video generation},
  author={Che, Haoxuan and He, Xuanhua and Liu, Quande and Jin, Cheng and Chen, Hao},
  journal={arXiv preprint arXiv:2411.00769},
  year={2024}
}

@misc{rae,
      title={Diffusion Transformers with Representation Autoencoders}, 
      author={Boyang Zheng and Nanye Ma and Shengbang Tong and Saining Xie},
      year={2025},
      eprint={2510.11690},
      archivePrefix={arXiv},
      primaryClass={cs.CV},
      url={https://arxiv.org/abs/2510.11690}, 
}

@article{su2024roformer,
  title={Roformer: Enhanced transformer with rotary position embedding},
  author={Su, Jianlin and Ahmed, Murtadha and Lu, Yu and Pan, Shengfeng and Bo, Wen and Liu, Yunfeng},
  journal={Neurocomputing},
  volume={568},
  pages={127063},
  year={2024},
  publisher={Elsevier}
}

@article{hu2023gaia,
  title={Gaia-1: A generative world model for autonomous driving},
  author={Hu, Anthony and Russell, Lloyd and Yeo, Hudson and Murez, Zak and Fedoseev, George and Kendall, Alex and Shotton, Jamie and Corrado, Gianluca},
  journal={arXiv preprint arXiv:2309.17080},
  year={2023}
}

@article{parkerholder2024genie2,
  title         = {Genie 2: A Large-Scale Foundation World Model},
  author        = {Jack Parker-Holder and Philip Ball and Jake Bruce and Vibhavari Dasagi and Kristian Holsheimer and Christos Kaplanis and Alexandre Moufarek and Guy Scully and Jeremy Shar and Jimmy Shi and Stephen Spencer and Jessica Yung and Michael Dennis and Sultan Kenjeyev and Shangbang Long and Vlad Mnih and Harris Chan and Maxime Gazeau and Bonnie Li and Fabio Pardo and Luyu Wang and Lei Zhang and Frederic Besse and Tim Harley and Anna Mitenkova and Jane Wang and Jeff Clune and Demis Hassabis and Raia Hadsell and Adrian Bolton and Satinder Singh and Tim Rockt{\"a}schel},
  year          = {2024},
  url           = {https://deepmind.google/discover/blog/genie-2-a-large-scale-foundation-world-model/}
}

@article{valevski2024diffusion,
  title={Diffusion models are real-time game engines},
  author={Valevski, Dani and Leviathan, Yaniv and Arar, Moab and Fruchter, Shlomi},
  journal={arXiv preprint arXiv:2408.14837},
  year={2024}
}

@article{xie2024progressive,
  title={Progressive autoregressive video diffusion models},
  author={Xie, Desai and Xu, Zhan and Hong, Yicong and Tan, Hao and Liu, Difan and Liu, Feng and Kaufman, Arie and Zhou, Yang},
  journal={arXiv preprint arXiv:2410.08151},
  year={2024}
}

@article{azzolini2025cosmos,
  title={Cosmos-reason1: From physical common sense to embodied reasoning},
  author={Azzolini, Alisson and Brandon, Hannah and Chattopadhyay, Prithvijit and Chen, Huayu and Chu, Jinju and Cui, Yin and Diamond, Jenna and Ding, Yifan and Ferroni, Francesco and Govindaraju, Rama and others},
  journal={arXiv preprint arXiv:2503.15558},
  year={2025}
}

@article{yu2025gamefactory,
  title={GameFactory: Creating New Games with Generative Interactive Videos},
  author={Yu, Jiwen and Qin, Yiran and Wang, Xintao and Wan, Pengfei and Zhang, Di and Liu, Xihui},
  journal={arXiv preprint arXiv:2501.08325},
  year={2025}
}

@inproceedings{peebles2023scalable,
  title={Scalable diffusion models with transformers},
  author={Peebles, William and Xie, Saining},
  booktitle={Proceedings of the IEEE/CVF international conference on computer vision},
  pages={4195--4205},
  year={2023}
}

@article{ha2018world,
  title={World models},
  author={Ha, David and Schmidhuber, J{\"u}rgen},
  journal={arXiv preprint arXiv:1803.10122},
  year={2018}
}

@article{oasis2024,
  author    = {Decart and Etched and Julian Quevedo and Quinn McIntyre and Spruce Campbell and Xinlei Chen and Robert Wachen},
  title     = {Oasis: A Universe in a Transformer},
  year      = {2024},
  url       = {https://oasis-model.github.io/}
}

@article{ding2024dollar,
  title={Dollar: Few-step video generation via distillation and latent reward optimization},
  author={Ding, Zihan and Jin, Chi and Liu, Difan and Zheng, Haitian and Singh, Krishna Kumar and Zhang, Qiang and Kang, Yan and Lin, Zhe and Liu, Yuchen},
  journal={arXiv preprint arXiv:2412.15689},
  year={2024}
}

@article{wang2023modelscope,
  title={Modelscope text-to-video technical report},
  author={Wang, Jiuniu and Yuan, Hangjie and Chen, Dayou and Zhang, Yingya and Wang, Xiang and Zhang, Shiwei},
  journal={arXiv preprint arXiv:2308.06571},
  year={2023}
}

@article{yang2024cogvideox,
  title={Cogvideox: Text-to-video diffusion models with an expert transformer},
  author={Yang, Zhuoyi and Teng, Jiayan and Zheng, Wendi and Ding, Ming and Huang, Shiyu and Xu, Jiazheng and Yang, Yuanming and Hong, Wenyi and Zhang, Xiaohan and Feng, Guanyu and others},
  journal={arXiv preprint arXiv:2408.06072},
  year={2024}
}

@article{singer2022make,
  title={Make-A-Video: Text-to-Video Generation without Text-Video Data},
  author={Singer, Uriel and Polyak, Adam and Nachmani, Eliya and Dahan, Guy and Shechtman, Eli and Hacohen, Haggai},
  journal={arXiv preprint arXiv:2209.14792},
  year={2022}
}

@article{hong2022cogvideo,
  title={CogVideo: Large-scale Pretraining for Text-to-Video Generation with Transformers},
  author={Hong, Yu and Wei, Jing and Liu, Xing and Wang, Xiaodi and Bai, Yutong and Li, Haitao and Zhang, Ming and Xu, Hao},
  journal={arXiv preprint arXiv:2205.15868},
  year={2022}
}

@article{ho2022video,
  title={Video Diffusion Models},
  author={Ho, Jonathan and Salimans, Tim and Gritsenko, Alexey and Chan, William and Norouzi, Mohammad and Fleet, David J.},
  journal={arXiv preprint arXiv:2204.03458},
  year={2022}
}

@inproceedings{chen2024videocrafter2,
  title={Videocrafter2: Overcoming data limitations for high-quality video diffusion models},
  author={Chen, Haoxin and Zhang, Yong and Cun, Xiaodong and Xia, Menghan and Wang, Xintao and Weng, Chao and Shan, Ying},
  booktitle={Proceedings of the IEEE/CVF Conference on Computer Vision and Pattern Recognition},
  pages={7310--7320},
  year={2024}
}

@INPROCEEDINGS{10377444,
  author={Esser, Patrick and Chiu, Johnathan and Atighehchian, Parmida and Granskog, Jonathan and Germanidis, Anastasis},
  booktitle={2023 IEEE/CVF International Conference on Computer Vision (ICCV)}, 
  title={Structure and Content-Guided Video Synthesis with Diffusion Models}, 
  year={2023},
  volume={},
  number={},
  pages={7312-7322},
  doi={10.1109/ICCV51070.2023.00675}}

@inproceedings{rombach2022high,
  title={High-resolution image synthesis with latent diffusion models},
  author={Rombach, Robin and Blattmann, Andreas and Lorenz, Dominik and Esser, Patrick and Ommer, Bj{\"o}rn},
  booktitle={Proceedings of the IEEE/CVF conference on computer vision and pattern recognition},
  pages={10684--10695},
  year={2022}
}

@misc{kingma2013auto,
  title={Auto-encoding variational bayes},
  author={Kingma, Diederik P and Welling, Max and others},
  year={2013},
  publisher={Banff, Canada}
}

@inproceedings{blattmann2023align,
  title={Align your latents: High-resolution video synthesis with latent diffusion models},
  author={Blattmann, Andreas and Rombach, Robin and Ling, Huan and Dockhorn, Tim and Kim, Seung Wook and Fidler, Sanja and Kreis, Karsten},
  booktitle={Proceedings of the IEEE/CVF Conference on Computer Vision and Pattern Recognition},
  pages={22563--22575},
  year={2023}
}

@article{wu2024ivideogpt,
  title={ivideogpt: Interactive videogpts are scalable world models},
  author={Wu, Jialong and Yin, Shaofeng and Feng, Ningya and He, Xu and Li, Dong and Hao, Jianye and Long, Mingsheng},
  journal={Advances in Neural Information Processing Systems},
  volume={37},
  pages={68082--68119},
  year={2024}
}

@article{chen2024diffusion,
  title={Diffusion forcing: Next-token prediction meets full-sequence diffusion},
  author={Chen, Boyuan and Mart{\'\i} Mons{\'o}, Diego and Du, Yilun and Simchowitz, Max and Tedrake, Russ and Sitzmann, Vincent},
  journal={Advances in Neural Information Processing Systems},
  volume={37},
  pages={24081--24125},
  year={2024}
}

@article{harvey2022flexible,
  title={Flexible Diffusion Modeling of Long Videos},
  author={Harvey, William and N{\o}rskov, S{\o}ren and K{\"o}lch, Niklas and Vogiatzis, George},
  journal={arXiv preprint arXiv:2205.11495},
  year={2022}
}

@article{blattmann2023stable,
  title={Stable video diffusion: Scaling latent video diffusion models to large datasets},
  author={Blattmann, Andreas and Dockhorn, Tim and Kulal, Sumith and Mendelevitch, Daniel and Kilian, Maciej and Lorenz, Dominik and Levi, Yam and English, Zion and Voleti, Vikram and Letts, Adam and others},
  journal={arXiv preprint arXiv:2311.15127},
  year={2023}
}

@article{guss2019minerl,
  title={Minerl: A large-scale dataset of minecraft demonstrations},
  author={Guss, William H and Houghton, Brandon and Topin, Nicholay and Wang, Phillip and Codel, Cayden and Veloso, Manuela and Salakhutdinov, Ruslan},
  journal={arXiv preprint arXiv:1907.13440},
  year={2019}
}

@INPROCEEDINGS{10657096,
  author={Huang, Ziqi and He, Yinan and Yu, Jiashuo and Zhang, Fan and Si, Chenyang and Jiang, Yuming and Zhang, Yuanhan and Wu, Tianxing and Jin, Qingyang and Chanpaisit, Nattapol and Wang, Yaohui and Chen, Xinyuan and Wang, Limin and Lin, Dahua and Qiao, Yu and Liu, Ziwei},
  booktitle={2024 IEEE/CVF Conference on Computer Vision and Pattern Recognition (CVPR)}, 
  title={VBench: Comprehensive Benchmark Suite for Video Generative Models}, 
  year={2024},
  volume={},
  number={},
  pages={21807-21818},
  doi={10.1109/CVPR52733.2024.02060}}

@misc{song2025historyguidedvideodiffusion,
      title={History-Guided Video Diffusion}, 
      author={Kiwhan Song and Boyuan Chen and Max Simchowitz and Yilun Du and Russ Tedrake and Vincent Sitzmann},
      year={2025},
      eprint={2502.06764},
      archivePrefix={arXiv},
      primaryClass={cs.LG},
      url={https://arxiv.org/abs/2502.06764}, 
}

@misc{zhang2025packinginputframecontext,
      title={Packing Input Frame Context in Next-Frame Prediction Models for Video Generation}, 
      author={Lvmin Zhang and Maneesh Agrawala},
      year={2025},
      eprint={2504.12626},
      archivePrefix={arXiv},
      primaryClass={cs.CV},
      url={https://arxiv.org/abs/2504.12626}, 
}

@misc{wang2025diffusedisperseimagegeneration,
      title={Diffuse and Disperse: Image Generation with Representation Regularization}, 
      author={Runqian Wang and Kaiming He},
      year={2025},
      eprint={2506.09027},
      archivePrefix={arXiv},
      primaryClass={cs.CV},
      url={https://arxiv.org/abs/2506.09027}, 
}

@inproceedings{yu2024repa,
    title={Representation Alignment for Generation: Training Diffusion Transformers Is Easier Than You Think},
    author={Sihyun Yu and Sangkyung Kwak and Huiwon Jang and Jongheon Jeong and Jonathan Huang and Jinwoo Shin and Saining Xie},
    year={2025},
    booktitle={International Conference on Learning Representations},
}

@article{balestriero2025lejepa,
  title={Lejepa: Provable and scalable self-supervised learning without the heuristics},
  author={Balestriero, Randall and LeCun, Yann},
  journal={arXiv preprint arXiv:2511.08544},
  year={2025}
}

@article{maes2026leworldmodel,
  title={Leworldmodel: Stable end-to-end joint-embedding predictive architecture from pixels},
  author={Maes, Lucas and Lidec, Quentin Le and Scieur, Damien and LeCun, Yann and Balestriero, Randall},
  journal={arXiv preprint arXiv:2603.19312},
  year={2026}
}

@inproceedings{edunov2019pre,
  title={Pre-trained language model representations for language generation},
  author={Edunov, Sergey and Baevski, Alexei and Auli, Michael},
  booktitle={Proceedings of the 2019 Conference of the North American Chapter of the Association for Computational Linguistics: Human Language Technologies, Volume 1 (Long and Short Papers)},
  pages={4052--4059},
  year={2019}
}

@inproceedings{NEURIPS2024_e304d374,
 author = {Li, Tianhong and Katabi, Dina and He, Kaiming},
 booktitle = {Advances in Neural Information Processing Systems},
 doi = {10.52202/079017-3985},
 editor = {A. Globerson and L. Mackey and D. Belgrave and A. Fan and U. Paquet and J. Tomczak and C. Zhang},
 pages = {125441--125468},
 publisher = {Curran Associates, Inc.},
 title = {Return of Unconditional Generation: A Self-supervised Representation Generation Method},
 url = {https://proceedings.neurips.cc/paper_files/paper/2024/file/e304d374c85e385eb217ed4a025b6b63-Paper-Conference.pdf},
 volume = {37},
 year = {2024}
}

@article{peale2025representative,
  title={Representative language generation},
  author={Peale, Charlotte and Raman, Vinod and Reingold, Omer},
  journal={arXiv preprint arXiv:2505.21819},
  year={2025}
}

@inproceedings{roy2007effective,
  title={The effective rank: A measure of effective dimensionality},
  author={Roy, Olivier and Vetterli, Martin},
  booktitle={2007 15th European signal processing conference},
  pages={606--610},
  year={2007},
  organization={IEEE}
}

@article{chen2026learning,
  title={Learning world models for interactive video generation},
  author={Chen, Taiye and Hu, Xun and Ding, Zihan and Jin, Chi},
  journal={Advances in Neural Information Processing Systems},
  volume={38},
  pages={154456--154483},
  year={2026}
}

@InProceedings{Yin_2025_CVPR,
    author    = {Yin, Tianwei and Zhang, Qiang and Zhang, Richard and Freeman, William T. and Durand, Fredo and Shechtman, Eli and Huang, Xun},
    title     = {From Slow Bidirectional to Fast Autoregressive Video Diffusion Models},
    booktitle = {Proceedings of the IEEE/CVF Conference on Computer Vision and Pattern Recognition (CVPR)},
    month     = {June},
    year      = {2025},
    pages     = {22963-22974}
}

@article{huang2026self,
  title={Self forcing: Bridging the train-test gap in autoregressive video diffusion},
  author={Huang, Xun and Li, Zhengqi and He, Guande and Zhou, Mingyuan and Shechtman, Eli},
  journal={Advances in Neural Information Processing Systems},
  volume={38},
  pages={167283--167308},
  year={2026}
}

@article{cui2025self,
  title={Self-forcing++: Towards minute-scale high-quality video generation},
  author={Cui, Justin and Wu, Jie and Li, Ming and Yang, Tao and Li, Xiaojie and Wang, Rui and Bai, Andrew and Ban, Yuanhao and Hsieh, Cho-Jui},
  journal={arXiv preprint arXiv:2510.02283},
  year={2025}
}

@misc{zhou2025dinowmworldmodelspretrained,
      title={DINO-WM: World Models on Pre-trained Visual Features enable Zero-shot Planning}, 
      author={Gaoyue Zhou and Hengkai Pan and Yann LeCun and Lerrel Pinto},
      year={2025},
      eprint={2411.04983},
      archivePrefix={arXiv},
      primaryClass={cs.RO},
      url={https://arxiv.org/abs/2411.04983}, 
}

@misc{bardes2024revisitingfeaturepredictionlearning,
      title={Revisiting Feature Prediction for Learning Visual Representations from Video}, 
      author={Adrien Bardes and Quentin Garrido and Jean Ponce and Xinlei Chen and Michael Rabbat and Yann LeCun and Mahmoud Assran and Nicolas Ballas},
      year={2024},
      eprint={2404.08471},
      archivePrefix={arXiv},
      primaryClass={cs.CV},
      url={https://arxiv.org/abs/2404.08471}, 
}

@misc{assran2025vjepa2selfsupervisedvideo,
      title={V-JEPA 2: Self-Supervised Video Models Enable Understanding, Prediction and Planning}, 
      author={Mido Assran and Adrien Bardes and David Fan and Quentin Garrido and Russell Howes and Mojtaba and Komeili and Matthew Muckley and Ammar Rizvi and Claire Roberts and Koustuv Sinha and Artem Zholus and Sergio Arnaud and Abha Gejji and Ada Martin and Francois Robert Hogan and Daniel Dugas and Piotr Bojanowski and Vasil Khalidov and Patrick Labatut and Francisco Massa and Marc Szafraniec and Kapil Krishnakumar and Yong Li and Xiaodong Ma and Sarath Chandar and Franziska Meier and Yann LeCun and Michael Rabbat and Nicolas Ballas},
      year={2025},
      eprint={2506.09985},
      archivePrefix={arXiv},
      primaryClass={cs.AI},
      url={https://arxiv.org/abs/2506.09985}, 
}

@article{genie3,
  title         = {Genie 3: A New Frontier for World Models},
  author        = {Philip J. Ball and Jakob Bauer and Frank Belletti and Bethanie Brownfield and Ariel Ephrat and Shlomi Fruchter and Agrim Gupta and Kristian Holsheimer and Aleksander Holynski and Jiri Hron and Christos Kaplanis and Marjorie Limont and Matt McGill and Yanko Oliveira and Jack Parker-Holder and Frank Perbet and Guy Scully and Jeremy Shar and Stephen Spencer and Omer Tov and Ruben Villegas and Emma Wang and Jessica Yung and Cip Baetu and Jordi Berbel and David Bridson and Jake Bruce and Gavin Buttimore and Sarah Chakera and Bilva Chandra and Paul Collins and Alex Cullum and Bogdan Damoc and Vibha Dasagi and Maxime Gazeau and Charles Gbadamosi and Woohyun Han and Ed Hirst and Ashyana Kachra and Lucie Kerley and Kristian Kjems and Eva Knoepfel and Vika Koriakin and Jessica Lo and Cong Lu and Zeb Mehring and Alex Moufarek and Henna Nandwani and Valeria Oliveira and Fabio Pardo and Jane Park and Andrew Pierson and Ben Poole and Helen Ran and Tim Salimans and Manuel Sanchez and Igor Saprykin and Amy Shen and Sailesh Sidhwani and Duncan Smith and Joe Stanton and Hamish Tomlinson and Dimple Vijaykumar and Luyu Wang and Piers Wingfield and Nat Wong and Keyang Xu and Christopher Yew and Nick Young and Vadim Zubov and Douglas Eck and Dumitru Erhan and Koray Kavukcuoglu and Demis Hassabis and Zoubin Gharamani and Raia Hadsell and A{\"a}ron van den Oord and Inbar Mosseri and Adrian Bolton and Satinder Singh and Tim Rockt{\"a}schel},
  year          = {2025},
  url           = {https://deepmind.google/blog/genie-3-a-new-frontier-for-world-models/}
}

@misc{nvidia2025cosmosworldfoundationmodel,
      title={Cosmos World Foundation Model Platform for Physical AI}, 
      author={NVIDIA and : and Niket Agarwal and Arslan Ali and Maciej Bala and Yogesh Balaji and Erik Barker and Tiffany Cai and Prithvijit Chattopadhyay and Yongxin Chen and Yin Cui and Yifan Ding and Daniel Dworakowski and Jiaojiao Fan and Michele Fenzi and Francesco Ferroni and Sanja Fidler and Dieter Fox and Songwei Ge and Yunhao Ge and Jinwei Gu and Siddharth Gururani and Ethan He and Jiahui Huang and Jacob Huffman and Pooya Jannaty and Jingyi Jin and Seung Wook Kim and Gergely Klár and Grace Lam and Shiyi Lan and Laura Leal-Taixe and Anqi Li and Zhaoshuo Li and Chen-Hsuan Lin and Tsung-Yi Lin and Huan Ling and Ming-Yu Liu and Xian Liu and Alice Luo and Qianli Ma and Hanzi Mao and Kaichun Mo and Arsalan Mousavian and Seungjun Nah and Sriharsha Niverty and David Page and Despoina Paschalidou and Zeeshan Patel and Lindsey Pavao and Morteza Ramezanali and Fitsum Reda and Xiaowei Ren and Vasanth Rao Naik Sabavat and Ed Schmerling and Stella Shi and Bartosz Stefaniak and Shitao Tang and Lyne Tchapmi and Przemek Tredak and Wei-Cheng Tseng and Jibin Varghese and Hao Wang and Haoxiang Wang and Heng Wang and Ting-Chun Wang and Fangyin Wei and Xinyue Wei and Jay Zhangjie Wu and Jiashu Xu and Wei Yang and Lin Yen-Chen and Xiaohui Zeng and Yu Zeng and Jing Zhang and Qinsheng Zhang and Yuxuan Zhang and Qingqing Zhao and Artur Zolkowski},
      year={2025},
      eprint={2501.03575},
      archivePrefix={arXiv},
      primaryClass={cs.CV},
      url={https://arxiv.org/abs/2501.03575}, 
}

@inproceedings{jin2025pyramidal,
  title={Pyramidal flow matching for efficient video generative modeling},
  author={Jin, Yang and Sun, Zhicheng and Li, Ningyuan and Xu, Kun and Jiang, Hao and Zhuang, Nan and Huang, Quzhe and Song, Yang and Mu, Yadong and Lin, Zhouchen},
  booktitle={International Conference on Learning Representations},
  volume={2025},
  pages={23378--23402},
  year={2025}
}

@inproceedings{qiu2024freenoise,
  title={Freenoise: Tuning-free longer video diffusion via noise rescheduling},
  author={Qiu, Haonan and Xia, Menghan and Zhang, Yong and He, Yingqing and Wang, Xintao and Shan, Ying and Liu, Ziwei},
  booktitle={International Conference on Learning Representations},
  volume={2024},
  pages={5260--5274},
  year={2024}
}

@inproceedings{3692070,
  author = {Ruhe, David and Heek, Jonathan and Salimans, Tim and Hoogeboom, Emiel},
  title = {Rolling diffusion models},
  year = {2024},
  publisher = {JMLR.org},
  booktitle = {Proceedings of the 41st International Conference on Machine Learning},
  articleno = {1744},
  numpages = {18},
  location = {Vienna, Austria},
  series = {ICML'24}
}

@inproceedings{khachatryan2023text2video,
  title={Text2video-zero: Text-to-image diffusion models are zero-shot video generators},
  author={Khachatryan, Levon and Movsisyan, Andranik and Tadevosyan, Vahram and Henschel, Roberto and Wang, Zhangyang and Navasardyan, Shant and Shi, Humphrey},
  booktitle={Proceedings of the IEEE/CVF International Conference on Computer Vision},
  pages={15954--15964},
  year={2023}
}

@inproceedings{li2026stable,
  title={Stable Video Infinity: Infinite-Length Video Generation with Error Recycling},
  author={Li, Wuyang and Pan, Wentao and Luan, Po-Chien and Gao, Yang and Alahi, Alexandre},
  booktitle={International Conference on Learning Representations 2025 (ICLR 2025)},
  year={2026}
}

@inproceedings{po2026bagger,
  title={Bagger: Backwards aggregation for mitigating drift in autoregressive video diffusion models},
  author={Po, Ryan and Chan, Eric Ryan and Chen, Changan and Wetzstein, Gordon},
  booktitle={Proceedings of the IEEE/CVF Conference on Computer Vision and Pattern Recognition},
  pages={43727--43739},
  year={2026}
}

@inproceedings{weng2024art,
  title={Art-v: Auto-regressive text-to-video generation with diffusion models},
  author={Weng, Wenming and Feng, Ruoyu and Wang, Yanhui and Dai, Qi and Wang, Chunyu and Yin, Dacheng and Zhao, Zhiyuan and Qiu, Kai and Bao, Jianmin and Yuan, Yuhui and others},
  booktitle={Proceedings of the IEEE/CVF Conference on Computer Vision and Pattern Recognition},
  pages={7395--7405},
  year={2024}
}

@misc{ren2025cosmosdrivedreamsscalablesyntheticdriving,
      title={Cosmos-Drive-Dreams: Scalable Synthetic Driving Data Generation with World Foundation Models}, 
      author={Xuanchi Ren and Yifan Lu and Tianshi Cao and Ruiyuan Gao and Shengyu Huang and Amirmojtaba Sabour and Tianchang Shen and Tobias Pfaff and Jay Zhangjie Wu and Runjian Chen and Seung Wook Kim and Jun Gao and Laura Leal-Taixe and Mike Chen and Sanja Fidler and Huan Ling},
      year={2025},
      eprint={2506.09042},
      archivePrefix={arXiv},
      primaryClass={cs.CV},
      url={https://arxiv.org/abs/2506.09042}, 
}

@article{ge2025,
    title={Genie Envisioner: A Unified World Foundation Platform for Robotic Manipulation},
    author={Yue Liao and Pengfei Zhou and Siyuan Huang and Donglin Yang and Shengcong Chen and Yuxin Jiang and Yue Hu and Jingbin Cai and Si Liu and Jianlan Luo and Liliang Chen and Shuicheng Yan and Maoqing Yao and Guanghui Ren},
    journal={arXiv preprint arXiv:2508.05635},
    year={2025}
}
\bibliographystyle{iclr2026_conference}


\end{document}